%% file: main.tex
\documentclass[10pt,twocolumn,letterpaper]{article}
\usepackage[pagenumbers]{cvpr}
\input{preamble}

\definecolor{cvprblue}{rgb}{0.21,0.49,0.74}
\usepackage[breaklinks,colorlinks,allcolors=cvprblue]{hyperref}
\hypersetup{
  pdftitle={Room for Supervision: Training-Free Dataset Distillation for Object Detection},
  pdfsubject={Dataset distillation for object detection},
  pdfkeywords={GRASP, dataset distillation, object detection}
}

\title{Room for Supervision:\\Training-Free Dataset Distillation for Object Detection}
\input{authors}
\date{}

\begin{document}
\maketitle
\input{sec/0_abstract}
\input{sec/1_intro}
\input{sec/2_related}
\input{sec/3_method}
\input{sec/4_experiments}
\input{sec/5_conclusion}

{\small
\bibliographystyle{ieeenat_fullname}
\bibliography{main}
}

\clearpage
\appendix
\setcounter{figure}{0}
\setcounter{table}{0}
\setcounter{equation}{0}
\setcounter{algorithm}{0}
\renewcommand{\thefigure}{S\arabic{figure}}
\renewcommand{\thetable}{S\arabic{table}}
\renewcommand{\theequation}{S\arabic{equation}}
\renewcommand{\thealgorithm}{S\arabic{algorithm}}
\renewcommand{\theHfigure}{supp.\arabic{figure}}
\renewcommand{\theHtable}{supp.\arabic{table}}
\renewcommand{\theHequation}{supp.\arabic{equation}}
\providecommand{\theHalgorithm}{\arabic{algorithm}}
\renewcommand{\theHalgorithm}{supp.\arabic{algorithm}}
\input{sec/s_construction}
\input{sec/s_sensitivity}
\input{sec/s_ipd_scaling}

\input{sec/s_qualitative}
\end{document}

%% file: preamble.tex
\usepackage{amsmath,amssymb,amsthm}
\usepackage{graphicx}
\usepackage{multirow}
\usepackage{algorithm}
\usepackage{algpseudocode}
\usepackage{tikz}
\usetikzlibrary{arrows.meta,positioning}
\newcommand{\method}{GRASP}
\newcommand{\odthree}{OD\textsuperscript{3}}
\newcommand{\odthreestar}{OD\textsuperscript{3,*}}

\newcommand{\ind}{\mathbf{1}}
\newtheorem{proposition}{Proposition}
\usepackage{needspace}
\BeforeBeginEnvironment{proposition}{\Needspace{7\baselineskip}}

%% file: authors.tex
\author{
Ziang Gan \qquad Qi Zhu \qquad Libao Zhang\thanks{Corresponding author: Libao Zhang (\href{mailto:libaozhang@bnu.edu.cn}{libaozhang@bnu.edu.cn}).}\\
School of Artificial Intelligence, Beijing Normal University\\
Beijing 100875, China\\
{\tt\small libaozhang@bnu.edu.cn}
}
\hypersetup{pdfauthor={Ziang Gan, Qi Zhu, Libao Zhang}}

%% file: sec/0_abstract.tex
\begin{abstract}
Dataset distillation (DD) reduces storage and training costs by compressing large datasets into small collections.
Most existing methods focus on image classification, where each image typically has a single category label.
In contrast, object detection requires accurate category and bounding box annotations for multiple objects, together with visual content consistent with those annotations.
Existing detection distillation pipelines rely on learned models and can alter the spatial relationships between objects and their backgrounds.
We introduce GRASP, which packs dense native supervision into limited pixels without network training or inference.
GRASP jointly searches over image selections and spatial layouts using convex allocation subproblems that penalize insufficient object resolution, excessive image compression, and padding that source context cannot fill.
During composition, each selected image is cropped around all annotated objects and expanded into surrounding context without reducing object scale.
The resized regions are then combined into distilled images, preserving the spatial layout within each region and retaining all original class labels and bounding box annotations.
We evaluate GRASP on MS COCO and PASCAL VOC with compression ratios ranging from 0.25\% to 2\%.
GRASP constructs the 1\% MS COCO dataset in under two minutes.
Under a common training protocol using hard labels, GRASP achieves the state of the art with both CNN and ViT backbones.
Code is available at \url{https://github.com/tade1s/GRASP}.
\end{abstract}

%% file: sec/1_intro.tex
\section{Introduction}
\label{sec:intro}

Dataset distillation (DD) compresses a large training set into a small collection of informative images that supports repeated model development at low storage and training cost~\cite{wang2018dataset}.
Optimization-based methods match gradients~\cite{zhao2021gradient,kim2022idc}, training trajectories~\cite{cazenavette2022trajectory,guo2024datm}, or feature distributions~\cite{zhao2023distribution}, although their dependence on particular networks can limit transfer to unseen architectures~\cite{cazenavette2023glad}.
Decoupled synthesis further scales condensation to large image collections~\cite{yin2023sre2l,yin2024cda}.
Generative approaches exploit learned latent spaces~\cite{cazenavette2023glad}, while diffusion-based methods synthesize or refine distilled images using learned generative priors~\cite{gu2024minimax,santiago2025mgd3,chen2025influence,gan2026scg,xia2026diver}.
Extending these approaches from classification to detection requires preserving the visual evidence associated with each category and bounding box across objects of different sizes.

\input{fig/overview}

Recent detection distillation methods address this challenge through model-based synthesis and screening.
DCOD inverts a pretrained detector~\cite{qi2024dcod}, UniDD uses task-guided diffusion followed by filtering and relabeling~\cite{qi2025unidd}, and \odthree{} assembles real instance crops screened by an observer detector~\cite{khatib2025od3}.
These operations can change object appearance, remove surrounding context, or alter the correspondence between visible content and annotations.
For example, an isolated pedestrian crop retains its category but loses its original relation to nearby vehicles and the road.
Relabeling can assign boxes to the resulting image, but cannot recover visual evidence already removed during construction.
Consequently, supervision can weaken even when class-and-box labels remain, motivating compression that directly preserves annotated source content.

Real-image composition offers a direct route to this goal, yet existing selection pipelines still rely on learned models.
RDED scores patches with a pretrained observer~\cite{sun2024rded}, FocusDD combines model-selected patches with a downsampled source view to retain scene information~\cite{hu2025focusdd}, and region-refinement methods preserve salient content while updating other pixels~\cite{tran2025nrr}.
For detection, \odthree{} conditions candidate acceptance on a pretrained observer detector~\cite{khatib2025od3}, while PCA selects images using training-derived pruning scores and relies on Faster R-CNN with AUM for its detection extension~\cite{xiao2026pca}.
Such model-dependent scoring couples data selection to the scoring network's preferences and incurs substantial construction cost: on MS COCO, \odthree{} reports approximately 4.7 hours on one RTX 4090, with observer screening accounting for most of the runtime~\cite{khatib2025od3}, and the 12-epoch detector training needed to obtain PCA's pruning scores alone takes approximately 25 hours on a single RTX 5090.

Images in the original training dataset already pair real object appearances and scene context with native category and bounding-box annotations, providing a direct basis for compression from existing visual content.
Combining more of these images brings more annotated objects into a compact collection, increasing supervision density at a fixed output image count and resolution.
However, each added image also introduces background pixels that compete with objects for the available space.
Retaining entire images can devote much of this space to background far from annotated objects, forcing greater downscaling and reducing the detail available for recognition and localization.
Cropping objects individually reduces this background cost but removes surrounding context and breaks their original spatial relationships.
Detection distillation should therefore allocate its limited pixels to more annotated objects while preserving sufficient object resolution and scene context for these annotations to provide effective supervision.

We present \method{} (Geometry-guided Resolution Allocation and Scene Packing), a training-free pipeline that constructs compact detection datasets from source images selected using native category and object-scale statistics (\cref{fig:overview}).
For each selected image, we extract a crop enclosing all its annotated objects and the background between them, resize the crop as a whole, and map its native categories and boxes onto an output canvas.
Joint search over source selection and canvas partitions seeks to retain as much effective supervision as possible within the prescribed annotation and pixel budgets.
Simply including more annotations is insufficient: selection can become dominated by tiny objects, packing can excessively shrink source content, and the category--scale distribution can drift from the original dataset.
We therefore conserve the prescribed annotation count and enforce category--scale coverage during selection.
For each fixed source subset and partition structure, a convex objective reduces object-resolution deficits, excessive per-image shrinkage, and padding that the original image cannot fill.
Context recovery then expands each crop into surrounding source pixels without reducing object scale, and its padding-free aspect-ratio interval directly guides the allocation itself.

We evaluate the exported images using hard class-and-box labels across convolutional and Transformer detectors.
Classification studies show that soft labels carry structured teacher information beyond category identities, so downstream performance reflects both image content and knowledge conveyed through supervision~\cite{qin2024label}.
Under some soft-label evaluation protocols, random image subsets perform comparably to distilled sets, obscuring differences in the quality of the distilled images themselves~\cite{dey2026hard}.
We therefore evaluate \method{} using its retained native annotations, without an online teacher or auxiliary feature losses, to directly assess the training value of the constructed data.

At the 0.25\% MS COCO budget, \method{} reaches 12.0 AP on Faster R-CNN-R50 and 15.6 AP on ViTDet-B, exceeding \odthree{} by 4.2 and 8.0 AP, respectively.
Both results also surpass Random, Uniform, and \odthree{} trained with four times as many images (\cref{tab:coco_main}).
The 1\% MS COCO dataset contains 60,000 native annotations and is generated in 114.27 seconds without network training or inference.

%% file: fig/overview.tex
\begin{figure*}[t]
  \centering
  \includegraphics[width=\linewidth]{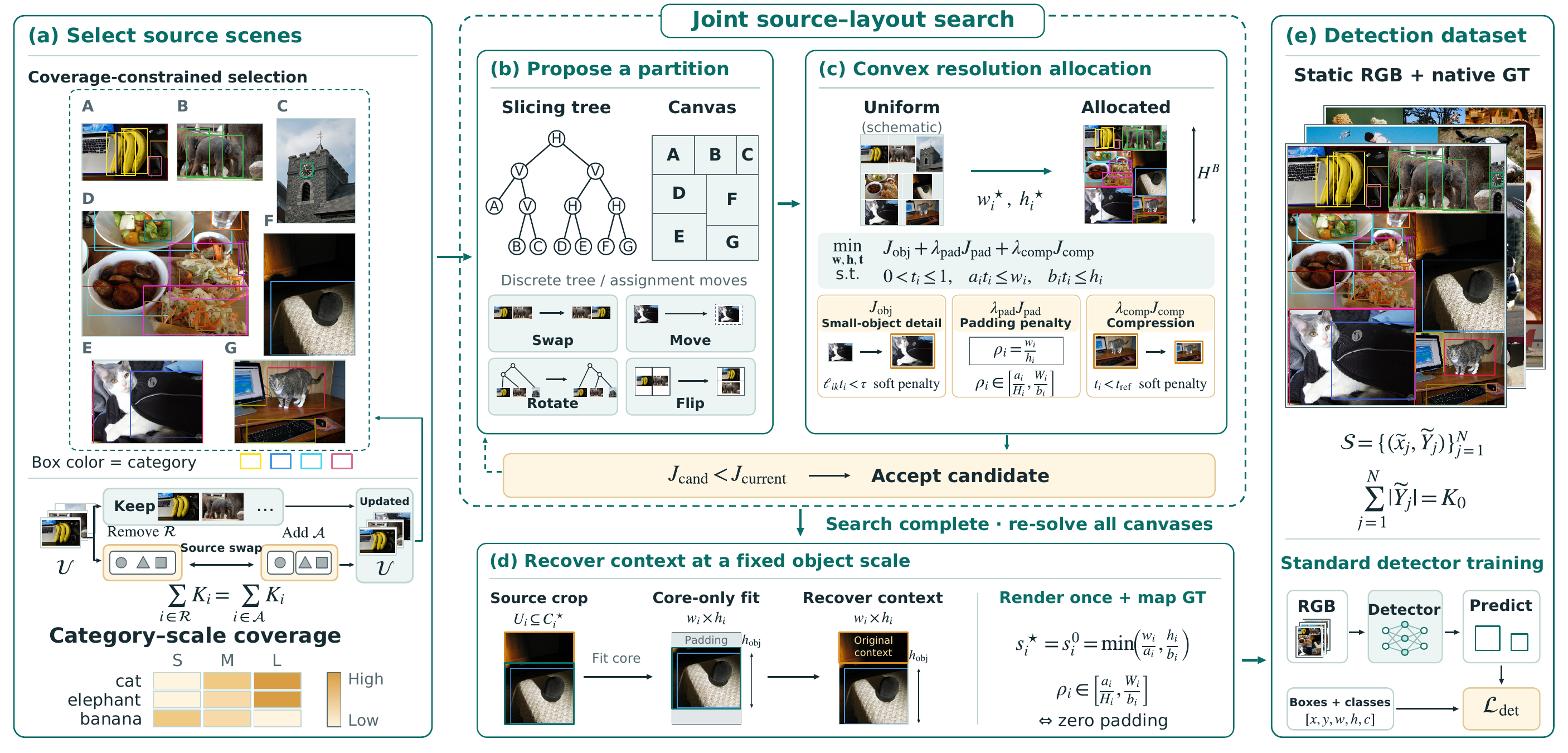}
  \caption{\textbf{Illustration of the \method{} framework.}
  \method{} selects real scenes, allocates resolution, and retains native supervision.
  (a) Source initialization and replacement satisfy the annotation-count and coverage constraints.
  (b) Local proposals change source assignments and binary partitions.
  (c) For each fixed structure, convex allocation balances object resolution, padding, and compression;
  feasible proposals are accepted only when the global objective decreases.
  (d) After search and a final allocation solve, context recovery preserves continuous object scale before rendering images and mapping boxes.
  (e) The resulting distilled dataset supports detector training with hard labels.}
  \label{fig:overview}
\end{figure*}

%% file: sec/2_related.tex
\section{Related Work}
\label{sec:related}

Dataset distillation synthesizes a compact set of training images that preserves generalization performance~\cite{wang2018dataset}.
Optimization-based methods align gradients~\cite{zhao2021gradient}, multi-step training trajectories~\cite{cazenavette2022trajectory}, or intermediate feature distributions~\cite{zhao2023distribution} computed through reference networks.
IDC uses multi-formation parameterization to represent more synthetic samples within a storage budget~\cite{kim2022idc}, while DATM aligns trajectory difficulty with the size of the distilled set~\cite{guo2024datm}.
SRe$^2$L separates model squeezing, image recovery, and relabeling for ImageNet-scale condensation~\cite{yin2023sre2l}; CDA uses a global-to-local synthesis curriculum to improve large-scale condensation~\cite{yin2024cda}.
To improve transfer across student architectures, GLaD parameterizes distilled samples in the latent space of a deep generative prior~\cite{cazenavette2023glad}.
Diffusion-based approaches use learned priors to generate or refine informative training samples~\cite{gu2024minimax,santiago2025mgd3,chen2025influence,gan2026scg,xia2026diver}.
Geometry-Aware Dataset Condensation selects a compact real subset by aligning distributions for diffusion model training~\cite{cui2026geometry}.
Real-image composition offers an alternative to pixel-space optimization: RDED uses observer-scored crops, fixed-grid packing, and teacher-generated labels~\cite{sun2024rded}.
PCA unifies pruning and distillation by composing easy real images under constrained augmentation and hard labels, using training-derived pruning scores~\cite{xiao2026pca}.
FocusDD combines ViT-selected patches with a downsampled source view to retain scene information~\cite{hu2025focusdd}.
NRR-DD combines salient patches and refines non-critical regions while preserving instance-specific detail~\cite{tran2025nrr}.

Object detection requires compact images to support simultaneous multi-instance localization and classification.
DCOD synthesizes detection data by inverting a pretrained detector with separate foreground and background stages~\cite{qi2024dcod}.
UniDD uses a task-specific proxy to guide diffusion sampling, then filters and relabels generated scenes for VOC and MS COCO~\cite{qi2025unidd}.
PCA's detection extension uses Faster R-CNN-C4 with ResNet-50 to collect AUM training dynamics for pruning before composing VOC images~\cite{xiao2026pca}.
\odthree{} assembles real instance regions, screens candidate layouts with a pretrained observer, and uses PKD feature distillation during downstream training~\cite{khatib2025od3}.
Mosaic mixes four source images during detector training to diversify scale and context, rather than producing a fixed compressed dataset~\cite{bochkovskiy2020yolov4}.
\method{} exports a reusable compact collection by retaining all valid annotations within each selected source region and allocating space from box sizes and available context, without model-based selection or pixel refinement.

%% file: sec/3_method.tex
\section{Method}
\label{sec:method}

\subsection{Problem Setting and Overview}
\label{sec:formulation}

Our goal is to construct a compact dataset $\mathcal S=\{(\widetilde x_j,\widetilde Y_j)\}_{j=1}^{N}$ containing $N$ images of resolution $W^B\times H^B$ and a total of $K_0$ annotations from a large original dataset $\mathcal D=\{(x_i,Y_i,W_i,H_i)\}_{i=1}^{D}$.
Here $x_i$ is an RGB image of size $W_i\times H_i$ with annotation set $Y_i$; $\widetilde x_j$ and $\widetilde Y_j$ are the output image and mapped annotations.
For an image with $K_i$ valid annotations, we write
\begin{equation}
\label{eq:annotations}
Y_i=\{(b_{ik},c_{ik},\mathrm{id}_{ik})\}_{k=1}^{K_i},\quad
b_{ik}=(x_{ik},y_{ik},\omega_{ik},\nu_{ik}).
\end{equation}
The $k$th annotation comprises a bounding box $b_{ik}$, category $c_{ik}$, and unique source annotation identifier $\mathrm{id}_{ik}$; $(x_{ik},y_{ik})$ is the box's top-left corner and $(\omega_{ik},\nu_{ik})$ its width and height in source pixels.
We select a subset of distinct source images, indexed by $\mathcal U=\{i:z_i=1\}$, where $z_i\in\{0,1\}$ indicates whether image $i$ is selected.
The output satisfies
\begin{equation}
\label{eq:budget}
|\mathcal S|=N,\qquad \sum_i z_iK_i=K_0,\qquad B=NW^BH^B.
\end{equation}
We fix pixel budget $B$ and annotation count $K_0$, then optimize the visual content supporting this supervision.
Following the images-per-dataset (IPD) convention of \odthree{}~\cite{khatib2025od3}, we specify the image budget by $N$ and report its nominal ratio $100N/D$ as IPD (\%) in the tables.
Each selected image contributes one rectangular region containing all its valid annotations.
Every canvas is nonempty, so the selected image count $M=|\mathcal U|\ge N$ is variable.

For each image with valid annotations ($K_i>0$), we define a core rectangle $U_i$ enclosing all its boxes, expanded on each side by a fraction $\eta$ of the envelope's width or height, rounded outward, and limited to the source image boundaries.
Its width and height are $a_i$ and $b_i$, and every crop used for image $i$ must contain this rectangle.
All annotated objects and the background between them remain in one region, so their relative positions are preserved by the common crop-and-resize mapping.
Images without valid annotations use the full image as $U_i$.

\paragraph{Pipeline overview.}
\Cref{fig:overview} illustrates the \method{} framework.
\method{} initializes a coverage-constrained subset (\cref{sec:selection}), jointly searches over sources and layouts via convex resolution allocation (\cref{sec:allocation}), and expands each core crop into surrounding source pixels at a fixed object scale before rendering (\cref{sec:context}).

\subsection{Coverage-Constrained Subset Initialization}
\label{sec:selection}

For object detection, object scale matters alongside category: detectors must recognize and localize objects across a wide range of sizes~\cite{lin2017fpn}.
We therefore seek an image subset containing exactly $K_0$ annotations whose joint category--scale distribution approximates that of the full dataset.
Coverage is evaluated in source coordinates so that selection feasibility remains invariant to subsequent resizing.
For each instance $(i,k)$, we compute its normalized scale $a^{\mathrm{src}}_{ik}$ and accumulate per-image category--scale counts $h_{i,c,b}$:
\begin{equation}
\begin{split}
a^{\mathrm{src}}_{ik}&=\sqrt{\omega_{ik}\nu_{ik}/(W_iH_i)},\\
h_{i,c,b}&=\sum_{k=1}^{K_i}\ind[c_{ik}=c,\ a^{\mathrm{src}}_{ik}\in I_b],
\end{split}
\end{equation}
where $c$ indexes categories, $b$ indexes three scale bins $\{I_b\}_{b=1}^3$ defined by source-instance tertiles, and $\ind[\cdot]$ is the indicator function.
Normalizing over the full dataset and the selected subset yields $p_{c,b}=\sum_i h_{i,c,b}/\sum_iK_i$ and $\widehat p_{c,b}=K_0^{-1}\sum_i z_i h_{i,c,b}$.
We bound their frequency-weighted $L_1$ discrepancy by
\begin{equation}
\label{eq:coverage}
D_{\mathrm{inst}}=\sum_{c,b}\gamma_{c,b}|\widehat p_{c,b}-p_{c,b}|
\le\varepsilon_{\mathrm{inst}}=0.10,
\end{equation}
where $\gamma_{c,b}=\widetilde\gamma_{c,b}/\sum_{c',b'}p_{c',b'}\widetilde\gamma_{c',b'}$ and $\widetilde\gamma_{c,b}=1/\sqrt{\max(p_{c,b},1/K_0)}$.
\Cref{eq:coverage} aligns the subset's joint category--scale distribution with the full dataset, giving greater weight to deviations in less frequent category--scale bins.
Per-category count tolerances prevent aggregate agreement from masking drift in rare categories.
Looser bounds regulate annotations per image, core-rectangle coverage $a_ib_i/(W_iH_i)$, and the fraction of images without annotations; Appendix~\ref{sec:s_coverage} specifies these constraints.

We initialize $\mathcal U$ with distinct source images containing exactly $K_0$ annotations and greedily refine source-distribution agreement to obtain a feasible starting subset.
We randomly permute the resulting $M=|\mathcal U|$ images and divide them into $N$ nonempty groups, each containing $\lfloor M/N\rfloor$ or $\lceil M/N\rceil$ images.
Each group is assigned to one canvas, and every selected image appears exactly once.
This subset and grouping initialize the joint optimization below, which constructs the initial partitions and maintains the same coverage constraints throughout search.

\subsection{Joint Image Selection and Resolution Allocation}
\label{sec:allocation}

Equal-sized regions can leave small objects with too few pixels while giving unnecessary space to images containing larger objects.
We optimize each region's width and height using the source box sizes, the rectangle enclosing all boxes, and the amount of surrounding image content available to fill unused space.
Given the initial subset and canvas groups from \cref{sec:selection}, we first construct and size an initial partition for each canvas.
An outer search then changes image selection and partition structure, while an inner convex program determines region dimensions for each fixed candidate.

\paragraph{Canvas partitions.}
Each canvas is represented by an axis-aligned binary partition tree: every internal node splits a rectangle in two, and each leaf is the region assigned to one source image.
For a fixed tree topology, split orientations, and leaf assignment, every node $v$ carries width and height variables $(w_v,h_v)$ anchored at root dimensions $(W^B,H^B)$.
A vertical split enforces $w_L+w_R=w_v$ and $h_L=h_R=h_v$, whereas a horizontal split enforces $h_L+h_R=h_v$ and $w_L=w_R=w_v$.
Placing child subtrees adjacently tiles the canvas without gaps or overlaps through purely linear equality constraints, eliminating nonlinear area-sum constraints.
Every leaf satisfies minimum dimension bounds $w_i,h_i\ge d_{\min}=16$ pixels.
To encourage composition, we cap the number $n_1$ of single-source canvases by $n_1\le\lfloor\max(0,2N-M)+\delta N\rfloor$, where $\delta$ controls the fractional allowance beyond the unavoidable minimum $\max(0,2N-M)$ (Appendix~\ref{sec:s_search}).

For each leaf $i$, we introduce an effective detail scale $t_i$ subject to
\begin{equation}
\label{eq:scale}
0<t_i\le1,\qquad a_it_i\le w_i,\qquad b_it_i\le h_i,
\end{equation}
where capping $t_i$ at $1$ reflects that upsampling beyond native resolution adds no new source detail.
Fitting the rectangle enclosing all source boxes into its assigned leaf may require downscaling the entire region.
This reduces the pixels supporting each object, potentially erasing appearance details and boundary cues needed for classification and localization, especially for small objects.
Letting $\ell_{ik}=\min(\omega_{ik},\nu_{ik})$ denote the native short side of box $b_{ik}$, we therefore penalize instances whose effective short side falls below the reference $\tau=16$ pixels:
\begin{equation}
\label{eq:obj}
J_{\mathrm{obj}}=\frac{1}{K_0}\sum_{i\in\mathcal U}\sum_{k=1}^{K_i}
\left[\log\frac{\tau}{\ell_{ik}t_i}\right]_+^2,
\end{equation}
where $[u]_+=\max(u,0)$ applies a smooth quadratic log-penalty below $\tau$ and zero penalty once $\ell_{ik}t_i\ge\tau$.

Because summing over instances weights densely annotated images more heavily, we pair $J_{\mathrm{obj}}$ with a per-image compression penalty, averaged equally over the $M_+=\sum_i z_i\ind[K_i>0]$ selected images with valid annotations:
\begin{equation}
\label{eq:comp}
J_{\mathrm{comp}}=\frac{1}{M_+}\sum_{i\in\mathcal U:K_i>0}
\left[\log\frac{t_{\mathrm{ref}}}{t_i}\right]_+^2,
\end{equation}
which penalizes excessive shrinkage of sparsely annotated images below reference scale $t_{\mathrm{ref}}=0.25$.

Expanding the $a_i\times b_i$ core within its $W_i\times H_i$ source image can fill a region of aspect ratio $\rho_i=w_i/h_i$ without padding or reduced object scale when (\cref{sec:context})
\begin{equation}
\label{eq:interval}
[\rho_i^{\min},\rho_i^{\max}]=[a_i/H_i,\ W_i/b_i].
\end{equation}
Outside this interval, the total thickness of padding strips that source context cannot eliminate is given by
\begin{equation}
\label{eq:padding}
\begin{split}
p_i(w_i,h_i)&=[w_i-\rho_i^{\max}h_i]_+\\
&\quad+[h_i-w_i/\rho_i^{\min}]_+,\\
J_{\mathrm{pad}}&=\frac{\sum_{i\in\mathcal U}p_i(w_i,h_i)}{N(W^B+H^B)}.
\end{split}
\end{equation}
Because $p_i(w_i,h_i)$ vanishes throughout $[\rho_i^{\min},\rho_i^{\max}]$, $J_{\mathrm{pad}}$ penalizes shape mismatch only beyond the context available in the source image while maintaining linearity inside each hinge for convex optimization.

For a fixed source subset and partition structure, we solve the continuous allocation subproblem
\begin{equation}
\label{eq:objective}
\begin{aligned}
\min_{\mathbf w,\mathbf h,\mathbf t}\quad&
J_{\mathrm{obj}}+\lambda_{\mathrm{pad}}J_{\mathrm{pad}}
+\lambda_{\mathrm{comp}}J_{\mathrm{comp}}\\
\text{s.t.}\quad&\text{root dimensions and binary split constraints},\\
&w_i,h_i\ge d_{\min},\quad\text{\cref{eq:scale}}.
\end{aligned}
\end{equation}

We use $\lambda_{\mathrm{pad}}=1$ and $\lambda_{\mathrm{comp}}=0.1$; Appendix~\ref{sec:s_search} provides solver settings.

\begin{proposition}[Convex allocation]
\label{prop:convex}
For a fixed source set and partition structure, \cref{eq:objective} is a convex optimization problem whenever $K_0,M_+>0$ and $\lambda_{\mathrm{pad}},\lambda_{\mathrm{comp}}\ge0$.
Under a fixed numerical lower bound $t_i\ge\epsilon_t>0$, its minimum is attained whenever the feasible set is nonempty.
\end{proposition}
The proof is provided in Appendix~\ref{sec:s_search}.

\paragraph{Initial layout.}
For each canvas group, we recursively bisect its ordered image list until each leaf contains one source image.
We construct two candidate trees, one with a vertical root split and one with a horizontal root split; deeper nodes split along the longer side of their provisional regions.
We solve \cref{eq:objective} for each candidate and retain the feasible partition with the lower objective value.
The resulting source assignments, trees, and optimized region dimensions form the starting state of the joint search.

\paragraph{Joint search.}
Starting from the initialized allocation, we replace selected indices $\mathcal R\subset\mathcal U$ with unselected indices $\mathcal A\subset\{1,\ldots,D\}\setminus\mathcal U$ satisfying
\begin{equation}
\label{eq:exchange}
\sum_{i\in\mathcal R}K_i=\sum_{i\in\mathcal A}K_i.
\end{equation}
These exchanges conserve $K_0$ while allowing the source count to change.
We also perturb canvas assignments and partition trees, re-solve only the affected canvases, and accept feasible proposals that decrease the globally normalized objective while maintaining source uniqueness and coverage constraints.
After a fixed budget of 500 proposals, we fix the selected sources and partition trees and re-solve all canvases using the final $M_+$ before context recovery and rendering.
Algorithm~\ref{alg:joint_search} in Appendix~\ref{sec:s_search} details the proposal distribution, normalization updates, and solver checks.

\subsection{Scale-Preserving Context Recovery}
\label{sec:context}

For a region $(w_i,h_i)$, fitting the core $U_i$ while preserving aspect ratio gives the continuous scale
\begin{equation}
s_i^0=\min(w_i/a_i,h_i/b_i).
\end{equation}
Unlike $s_i^0$, the effective detail scale is capped at $t_i^\star=\min(1,s_i^0)$.
We recover context along the non-limiting axis by expanding the crop to
\begin{equation}
\label{eq:context}
(W_i^{c\star},H_i^{c\star})=
\begin{cases}
(\min(W_i,\rho_i b_i),b_i),&\rho_i\ge a_i/b_i,\\
(a_i,\min(H_i,a_i/\rho_i)),&\rho_i<a_i/b_i,
\end{cases}
\end{equation}
centered on $U_i$ and clamped to $[0,W_i]\times[0,H_i]$ so that the resulting crop $C_i^\star$ always encloses $U_i$ (images without annotations retain the full source image).

This preserves $\min(w_i/W_i^{c\star},h_i/H_i^{c\star})=s_i^0$ and maximizes recovered context at that scale.
Padding vanishes when $\rho_i\in[\rho_i^{\min},\rho_i^{\max}]$; otherwise, its remaining strip thickness is exactly the penalty in \cref{eq:padding}.
Appendix~\ref{sec:s_rendering} gives the geometric derivation and integer implementation.

\paragraph{Rendering and labels.}
We round partitions to shared integer cuts, resample each recovered crop $C_i^\star$ once, and transform all native boxes using the realized integer crop, per-axis resize, and canvas translation (Appendix~\ref{sec:s_rendering}).
Preserving all native categories and boxes ($\sum_{j=1}^N|\widetilde Y_j|=K_0$), \method{} exports static RGB images and standard COCO-format annotations ready for off-the-shelf detector training.

%% file: sec/4_experiments.tex
\section{Experiments}
\label{sec:exp}

\setcounter{dbltopnumber}{1}

\begin{table*}[t]
\centering
\caption{\textbf{\method{} improves detection across budgets and architectures on MS COCO.} We report val2017 AP, AP$_{50}$, and AP$_{75}$ ($\uparrow$) after 264 epochs; \method{} results are mean$\pm$std over four evaluation seeds. $^*$ denotes hard-label evaluation of the released \odthree{} data~\cite{khatib2025od3}. $^\dagger$ denotes PCA~\cite{xiao2026pca} with single-tile cropping enabled for Faster R-CNN and disabled for ViTDet (the better tested settings); native detector augmentations remain enabled. \method{} uses 15K/30K/60K native annotations at 0.25\%/0.5\%/1\% IPD, respectively, and reuses each dataset across detectors. Bold: best within each detector--budget group.}
\label{tab:coco_main}
\small
\setlength{\tabcolsep}{4pt}
\renewcommand{\arraystretch}{1.08}
\begin{tabular*}{\textwidth}{@{\extracolsep{\fill}}clcccccc@{}}
\toprule
 & & \multicolumn{3}{c}{\textbf{Faster R-CNN-R50}} & \multicolumn{3}{c}{\textbf{ViTDet-B/16}} \\
\cmidrule(lr){3-5}\cmidrule(lr){6-8}
IPD (\%) & Method & AP & AP$_{50}$ & AP$_{75}$ & AP & AP$_{50}$ & AP$_{75}$ \\
\midrule
\multirow{5}{*}{0.25} & Random & 3.7 & 9.8 & 1.8 & 3.4 & 7.5 & 2.6 \\
 & Uniform & 3.8 & 9.9 & 1.8 & 3.5 & 7.6 & 2.6 \\
 & \odthreestar{}~\cite{khatib2025od3} & 7.8 & 16.0 & 6.8 & 7.6 & 15.4 & 6.8 \\
 & PCA$^\dagger$~\cite{xiao2026pca} & 10.6 & 19.8 & 10.0 & 13.0 & 22.1 & 13.3 \\
 & \method{} (ours) & \textbf{12.0$\pm$0.1} & \textbf{24.1$\pm$0.2} & \textbf{11.0$\pm$0.2} & \textbf{15.6$\pm$0.1} & \textbf{28.7$\pm$0.3} & \textbf{15.3$\pm$0.2} \\
\midrule
\multirow{5}{*}{0.50} & Random & 7.3 & 14.8 & 6.3 & 7.3 & 14.1 & 6.6 \\
 & Uniform & 7.4 & 14.9 & 6.4 & 7.4 & 14.3 & 6.7 \\
 & \odthreestar{}~\cite{khatib2025od3} & 9.5 & 18.6 & 8.8 & 11.0 & 20.8 & 10.6 \\
 & PCA$^\dagger$~\cite{xiao2026pca} & 13.3 & 24.0 & 13.3 & 16.3 & 26.7 & 17.1 \\
 & \method{} (ours) & \textbf{15.2$\pm$0.1} & \textbf{29.1$\pm$0.2} & \textbf{14.3$\pm$0.2} & \textbf{20.4$\pm$0.2} & \textbf{35.2$\pm$0.2} & \textbf{21.2$\pm$0.3} \\
\midrule
\multirow{5}{*}{1.00} & Random & 10.5 & 19.9 & 9.9 & 12.5 & 21.9 & 12.5 \\
 & Uniform & 10.6 & 20.1 & 9.9 & 12.6 & 22.0 & 12.5 \\
 & \odthreestar{}~\cite{khatib2025od3} & 11.6 & 22.1 & 11.1 & 12.9 & 24.0 & 12.6 \\
 & PCA$^\dagger$~\cite{xiao2026pca} & 16.1 & 27.8 & 16.7 & 19.2 & 30.5 & 20.6 \\
 & \method{} (ours) & \textbf{18.5$\pm$0.2} & \textbf{33.4$\pm$0.3} & \textbf{18.5$\pm$0.2} & \textbf{24.8$\pm$0.1} & \textbf{40.4$\pm$0.2} & \textbf{26.2$\pm$0.2} \\
\midrule
100 & Full Dataset & 39.8 & 60.1 & 43.3 & 51.6 & 72.1 & 56.6 \\
\bottomrule
\end{tabular*}
\end{table*}

\paragraph{Experimental Setup.}
We evaluate on MS COCO (train2017/val2017)~\cite{lin2014coco} and PASCAL VOC, following the image budgets and canvas resolutions of \odthree{}~\cite{khatib2025od3}.
For VOC, we use VOC2007+2012 trainval for construction and VOC2007 test for evaluation, as in \odthree{}.
For COCO, IPD values of 0.25\%/0.5\%/1\% correspond to $N=297/593/1{,}185$ images and $K_0=15\text{K}/30\text{K}/60\text{K}$ native annotations; for VOC, 0.5\%/1\%/2\% correspond to $N=83/166/332$ images and $K_0=4\text{K}/8\text{K}/16\text{K}$ annotations.
Canvas sizes are $484\times578$ for MS COCO and $375\times500$ for PASCAL VOC, matching the average width and height of their full training sets.
We set the per-side crop margin to $\eta=0.1$ and the single-source canvas allowance to $\delta=0.05$, and reuse each compressed dataset across Faster R-CNN-R50~\cite{ren2015faster,he2016resnet} and ViTDet-B/16~\cite{li2022vitdet} with their native augmentations.
Both detectors are trained for 264 epochs on COCO and VOC using an RTX 5090 GPU, a Ryzen 9 9950X CPU, and 96\,GB RAM.
We report final-epoch COCO AP, AP$_{50}$, and AP$_{75}$~\cite{lin2014coco} (mean$\pm$std over four evaluation seeds for \method{}) and all-points AP$_{50}$ on VOC following \odthree{}~\cite{khatib2025od3}.

\subsection{Comparison with the State of the Art}
\label{sec:sota}
We compare \method{} against \odthree{} (ICLR 2026)~\cite{khatib2025od3} and PCA (ICML 2026)~\cite{xiao2026pca}, along with Random and Uniform, the two strongest subset-selection baselines in \odthree{}'s evaluation.
Random samples source images uniformly at random, whereas Uniform balances selection across categories; both retain native annotations.
All methods are evaluated under the same hard-label protocol at matched image budgets.
For \odthree{}, we train detectors on its released COCO RGB images and observer-generated hard annotations without downstream teacher distillation (denoted \odthreestar{} in the tables).
PCA composes $2\times2$ grids and uses \texttt{PCASingleTile} to randomly select one tile before detector augmentations.
We report the better tested setting per detector on both datasets: tile cropping enabled for Faster R-CNN and disabled for ViTDet, retaining native detector augmentations.
At 0.5\% VOC, disabling \texttt{PCASingleTile} lowers Faster R-CNN AP$_{50}$ from 37.94 to 34.52 but raises ViTDet AP$_{50}$ from 5.45 to 33.92.
Appendix~\ref{sec:s_qualitative} visualizes distilled RGB canvases from \method{}, PCA, and \odthree{}.

\begin{table*}[t]
\centering
\caption{\textbf{Detection performance on PASCAL VOC.} We report all-points AP$_{50}$ ($\uparrow$) for Faster R-CNN-R50 and ViTDet-B/16. $^*$ denotes hard-label evaluation of \odthree{}~\cite{khatib2025od3}. $^\dagger$ denotes PCA~\cite{xiao2026pca} with single-tile cropping enabled for Faster R-CNN and disabled for ViTDet (the better tested settings); native detector augmentations remain enabled. Bold: best within each detector--budget group.}
\label{tab:voc_main}
\small
\setlength{\tabcolsep}{4.5pt}
\renewcommand{\arraystretch}{1.08}
\begin{tabular*}{\textwidth}{@{\extracolsep{\fill}}crrrrrrrrrr@{}}
\toprule
 & \multicolumn{5}{c}{\textbf{Faster R-CNN-R50}} & \multicolumn{5}{c}{\textbf{ViTDet-B/16}} \\
\cmidrule(lr){2-6}\cmidrule(lr){7-11}
IPD (\%) & Random & Uniform & PCA$^\dagger$ & \odthreestar{} & \method{} & Random & Uniform & PCA$^\dagger$ & \odthreestar{} & \method{} \\
\midrule
0.50 & 19.36 & 19.41 & 37.94 & 30.72 & \textbf{41.91} & 3.14 & 3.18 & 33.92 & 19.00 & \textbf{44.71} \\
1.00 & 32.75 & 32.95 & 46.97 & 34.87 & \textbf{47.63} & 3.28 & 3.41 & 43.38 & 25.40 & \textbf{58.82} \\
2.00 & 42.04 & 42.14 & 53.59 & 39.95 & \textbf{54.59} & 5.72 & 5.82 & 52.54 & 32.25 & \textbf{66.86} \\
\midrule
100 & \multicolumn{5}{c}{80.35} & \multicolumn{5}{c}{85.20} \\
\bottomrule
\end{tabular*}
\end{table*}

\paragraph{Main results.}
\method{} achieves the highest overall AP at every image budget for both detectors (\cref{tab:coco_main}).
Compared with PCA, its AP gains are 1.4/1.9/2.4 for Faster R-CNN and 2.6/4.1/5.6 for ViTDet at 0.25\%/0.5\%/1\%, respectively.
Relative to \odthree{}, Faster R-CNN-R50 improves by 4.2, 5.7, and 6.9 AP at the 0.25\%, 0.5\%, and 1\% budgets, respectively, while ViTDet-B improves by 8.0, 9.4, and 11.9 AP.
With only 297 images, \method{} also exceeds Random, Uniform, and \odthree{} trained on four times as many images (1,185 images) across both architectures.

\paragraph{PASCAL VOC.}
\method{} achieves the highest AP$_{50}$ across all three budgets and both detectors (\cref{tab:voc_main}), exceeding PCA by 3.97/0.66/1.00 points for Faster R-CNN and by 10.79/15.44/14.32 points for ViTDet.
Relative to \odthree{}, the gains reach 11.19/12.76/14.64 and 25.71/33.42/34.61 points, respectively.

\paragraph{Construction efficiency.}
\method{} constructs the 1\% COCO dataset with 60,000 annotations in 114.27 seconds on the Ryzen 9 9950X CPU, without neural network training or inference.
In contrast, \odthree{} reports approximately 4.7 hours for COCO construction on an RTX 4090, dominated by observer screening~\cite{khatib2025od3}.
PCA requires AUM-based pruning scores~\cite{xiao2026pca}; in our COCO reproduction, obtaining these scores alone takes approximately 25 hours of detector training for 12 epochs on a single RTX 5090 before image composition.

\subsection{Ablation Studies}
\label{sec:ablations}
All ablations use Faster R-CNN-R50 on MS COCO and report epoch-264 performance.
Unless otherwise specified, we evaluate at the 0.25\% COCO budget ($N=297$, $K_0=15\text{K}$), where the full method achieves 12.0 AP and 7.2 AP$_s$.

\begin{figure}[t]
\centering
\includegraphics[width=\columnwidth]{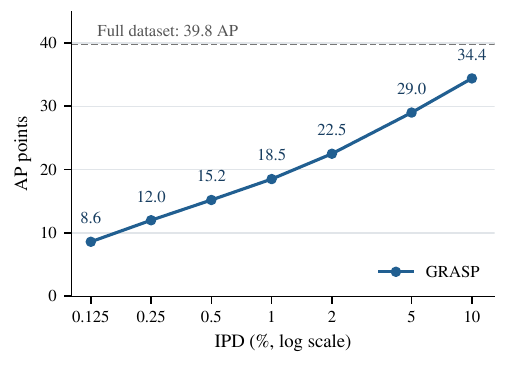}
\caption{\textbf{\method{} benefits from larger image budgets.} Points show epoch-264 AP across a logarithmic IPD axis as image and annotation budgets grow together. The dashed line marks the full-dataset reference.}
\label{fig:ipd_scaling}
\end{figure}

\paragraph{Scaling with the image budget.}
\label{sec:ipd_scaling}
\method{} scales consistently across seven COCO image budgets from 0.125\% to 10\% IPD with Faster R-CNN-R50 (\cref{fig:ipd_scaling}).
We fix the canvas resolution at $484\times578$ and scale the image and annotation budgets together, maintaining approximately 50.6 annotations per canvas.
AP rises steadily from 8.6 with 149 images at 0.125\% to 22.5, 29.0, and 34.4 at 2\%, 5\%, and 10\%, respectively, reaching within 5.4 AP of the full-dataset reference (39.8 AP).
Table~\ref{tab:ipd_scaling} reports image counts, annotation budgets, and all six detection metrics.

\begin{table}[t]
\centering
\caption{\textbf{Ablation of objective terms throughout construction.} $\checkmark$/$\times$: enabled/disabled; last column: full method. AP points ($\uparrow$).}
\label{tab:loss_ablation}
\small
\setlength{\tabcolsep}{3.5pt}
\renewcommand{\arraystretch}{1.08}
\begin{tabular*}{\columnwidth}{@{\extracolsep{\fill}}lcccccccc@{}}
\toprule
$J_{\mathrm{obj}}$ & $\times$ & $\checkmark$ & $\times$ & $\times$ & $\checkmark$ & $\checkmark$ & $\times$ & $\checkmark$ \\
$J_{\mathrm{pad}}$ & $\times$ & $\times$ & $\checkmark$ & $\times$ & $\checkmark$ & $\times$ & $\checkmark$ & $\checkmark$ \\
$J_{\mathrm{comp}}$ & $\times$ & $\times$ & $\times$ & $\checkmark$ & $\times$ & $\checkmark$ & $\checkmark$ & $\checkmark$ \\
\midrule
AP & 9.7 & 10.9 & 11.8 & 11.2 & 11.9 & 11.3 & 11.5 & \textbf{12.0} \\
AP$_s$ & 5.0 & 6.6 & 6.3 & 6.1 & 6.6 & 7.0 & 6.6 & \textbf{7.2} \\
\bottomrule
\end{tabular*}
\end{table}

\paragraph{Objective terms.}
\Cref{tab:loss_ablation} ablates the three terms in \cref{eq:objective}---object resolution ($J_{\mathrm{obj}}$), unrecoverable padding ($J_{\mathrm{pad}}$), and image compression ($J_{\mathrm{comp}}$)---using weights $1/1/0.1$ when enabled.
Starting from the same coverage-constrained initialization, each active variant applies its objective throughout initial tree selection, joint source--layout search, and final refinement, whereas the all-zero baseline partitions randomly grouped sources without objective-driven search.
Each term individually improves over the all-zero baseline (9.7 AP, 5.0 AP$_s$): $J_{\mathrm{pad}}$ reduces wasted canvas space, while $J_{\mathrm{obj}}$ and $J_{\mathrm{comp}}$ preserve instance and scene scale.
Combining all three terms achieves the highest overall AP (12.0) and small-object AP$_s$ (7.2).

\begin{table}[tb]
\centering
\caption{\textbf{Construction components and coverage strategies.} Settings (a)--(f) are defined in the text; (a): full method. AP points ($\uparrow$).}
\label{tab:component_ablation}
\small
\setlength{\tabcolsep}{4.5pt}
\renewcommand{\arraystretch}{1.08}
\begin{tabular*}{\columnwidth}{@{\extracolsep{\fill}}lcccccc@{}}
\toprule
Setting & (a) & (b) & (c) & (d) & (e) & (f) \\
\midrule
AP & \textbf{12.0} & 10.9 & 11.3 & 11.4 & 11.5 & 11.5 \\
AP$_s$ & \textbf{7.2} & 5.8 & 6.3 & 6.4 & 6.6 & 6.0 \\
\bottomrule
\end{tabular*}
\end{table}

\paragraph{Components and coverage.}
\Cref{tab:component_ablation} evaluates six construction variants:
\textbf{(a)} the full \method{} method combining coverage initialization, joint source--layout search, convex allocation, and context recovery;
\textbf{(b)} simple packing, which retains coverage initialization and context recovery but uses leaf-count-proportional areas and heuristic long-axis bisection without joint search or convex allocation;
\textbf{(c)} class-only coverage, which removes the joint category--scale constraint ($D_{\mathrm{inst}}$) while retaining per-class tolerances and image-level bounds;
\textbf{(d)} no coverage, which starts from a random subset at $K_0$ annotations and runs joint search without distribution constraints;
\textbf{(e)} no context recovery, which fixes the full method's selected sources, trees, and cuts and disables only context expansion during rendering (\cref{sec:context}); and
\textbf{(f)} equal-area allocation, which fixes the full method's sources and trees, assigns region areas proportional to subtree leaf counts, and retains context recovery.
Replacing joint search and convex allocation with simple packing in (b) lowers AP/AP$_s$ by 1.1/1.4 points.
Relaxing distribution alignment to class-only (c) or no coverage (d) reduces AP/AP$_s$ by 0.7/0.9 and 0.6/0.8 points, respectively.
Holding the final sources and trees fixed, omitting context recovery (e) reduces AP/AP$_s$ by 0.5/0.6 points, while equal-area allocation (f) reduces AP/AP$_s$ by 0.5/1.2 points, confirming the value of convex allocation for small objects.

\paragraph{Geometry audit.}
Geometric statistics on the exported 0.25\% COCO datasets directly explain these downstream gains.
Compared with equal-area allocation on identical sources and trees, \method{} reduces the fraction of exported boxes with short sides below 16 pixels from 50.58\% to 44.24\% and raises their fifth-percentile short side from 2.63 to 3.55 pixels.
Adding $J_{\mathrm{comp}}$ raises the first-percentile realized scale of sparsely annotated sources (1--3 instances) from 0.0889 to 0.1411 on the same selected sources.
With sources and layout fixed, context recovery reduces padded canvas area from 14.31\% to 1.85\% while leaving the sub-16-pixel box fraction unchanged.
These measurements confirm that each component directly improves object scale and spatial utilization.

\begin{figure}[t]
\centering
\includegraphics[width=\columnwidth]{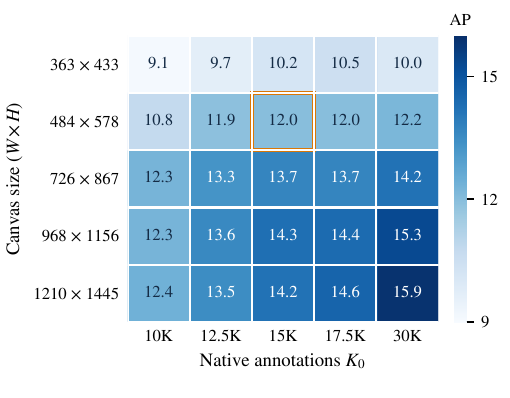}
\caption{\textbf{Larger canvases support more effective supervision.} COCO, 297 images, Faster R-CNN-R50, epoch 264. Each cell reports AP across canvas width$\times$height and annotation budget $K_0$. Orange outlines the default $484\times578$, $K_0=15$K setting. Increasing annotations helps on larger canvases but hurts at the highest annotation budget on the smallest canvas.}
\label{fig:canvas_k}
\end{figure}

\paragraph{Canvas size and annotation budget.}
\Cref{fig:canvas_k} evaluates five canvas resolutions across five annotation budgets with 297 images and the full pipeline.
Whereas \odthree{} reports lower AP when enlarging its canvas beyond $484\times578$ (Table~10 in~\cite{khatib2025od3}), \method{} benefits from the additional pixel capacity: at $K_0=15\text{K}$, AP increases from 12.0 at $484\times578$ to 14.3 at $968\times1156$ and remains at 14.2 at $1210\times1445$.
At $K_0=30\text{K}$, enlarging the canvas from $484\times578$ to $1210\times1445$ raises AP from 12.2 to 15.9, as larger canvases preserve finer visual support for each annotation.
The benefit of denser supervision likewise depends on available pixels: increasing $K_0$ from 17.5K to 30K raises AP from 14.6 to 15.9 on the largest canvas, but reduces AP from 10.5 to 10.0 on the smallest $363\times433$ canvas.
Together, these results show that additional annotations improve learning when objects retain sufficient visual detail, whereas excessive compression offsets that benefit; Appendix~\ref{sec:s_k_sweep} further examines this tradeoff across a wider fixed-resolution $K_0$ sweep.

%% file: sec/5_conclusion.tex
\section{Conclusion}
\label{sec:conclusion}

We presented \method{}, a training-free framework for object detection dataset distillation that allocates limited pixels to dense native supervision.
Coverage-constrained source selection and convex resolution allocation balance object detail, image compression, and unrecoverable padding, while scale-preserving context recovery retains surrounding visual evidence.
The resulting images preserve native class-and-box labels and can be reused across detector architectures without a construction-stage model.
Experiments on MS COCO and PASCAL VOC demonstrate state-of-the-art hard-label detection performance across budgets and backbones, with the 1\% COCO dataset constructed in 114.27 seconds.
These results support geometry-guided preservation of source content as an effective basis for detection distillation; extending the geometric objective to account for visual ambiguity remains an open direction.

\clearpage
\section*{LLM Usage Statement}
We used OpenAI's ChatGPT and Codex to assist with language polishing, coding, and internal manuscript review.
The authors are responsible for the final content.

%% file: sec/s_construction.tex
\section{Source Coverage and Initialization}
\label{sec:s_coverage}

We detail coverage, search, and rendering using the notation of Sec.~\ref{sec:method}.
For an image with valid annotations ($K_i>0$), the core rectangle $U_i=[u_i^-,u_i^+]\times[v_i^-,v_i^+]$ encloses all boxes with a margin of $\eta=0.1$ on each side, rounded outward and limited to $[0,W_i]\times[0,H_i]$.
Zero-instance images ($K_i=0$) set $U_i=[0,W_i]\times[0,H_i]$, and every valid source annotation is preserved throughout selection and rendering.
Valid annotations have a recognized category, are neither crowd nor ignore, have positive area and width and height of at least one pixel, and lie within the source image.

\paragraph{Category and scale tolerances.}
To prevent frequent categories from dominating the category--scale discrepancy $D_{\mathrm{inst}}$, we define inverse-square-root weights
\begin{equation}
\begin{split}
\widetilde\gamma_{c,b}&=\frac{1}{\sqrt{\max(p_{c,b},1/K_0)}},\\
\gamma_{c,b}&=\frac{\widetilde\gamma_{c,b}}
{\sum_{c',b'}p_{c',b'}\widetilde\gamma_{c',b'}},
\end{split}
\end{equation}
and enforce $D_{\mathrm{inst}}\le\varepsilon_{\mathrm{inst}}=0.10$.
For each category $c$, let actual and expected instance counts be
\begin{equation}
n_c=\sum_{i,b}z_i h_{i,c,b},\qquad \mu_c=K_0\sum_b p_{c,b}.
\end{equation}
We bound per-category drift via $|n_c-\mu_c|\le\Delta_c$, with
\begin{equation}
\Delta_c=\max\left(|n_c^{\mathrm{init}}-\mu_c|,
\left\lceil2\sqrt{\mu_c}+3\right\rceil\right),
\end{equation}
where $n_c^{\mathrm{init}}$ denotes the category count of a fixed exact-$K_0$ random reference draw established prior to optimization.

\paragraph{Image-level coverage statistics.}
Annotations per image are grouped into six bins: $0$, $1$, $2$--$3$, $4$--$7$, $8$--$15$, and $16+$.
Core-rectangle area coverage $v_i=a_ib_i/(W_iH_i)$ is grouped into four equal-width intervals: $[0,0.25)$, $[0.25,0.50)$, $[0.50,0.75)$, and $[0.75,1]$.
Normalizing selected histograms by the variable image count $M$ and source histograms by the dataset size $D$, we enforce
\begin{equation}
\begin{split}
\sum_t|\widehat p_t^{\mathrm{density}}-p_t^{\mathrm{density}}|&\le0.50,\\
\sum_r|\widehat p_r^{\mathrm{crop}}-p_r^{\mathrm{crop}}|&\le0.25,
\end{split}
\end{equation}
while bounding the absolute deviation in the fraction of images without annotations ($K_i=0$) by $0.02$.
Because all statistics depend solely on source annotations and $U_i$, source coverage feasibility remains invariant to canvas layout and context expansion.
Search initializes from a feasible subset of distinct source images containing exactly $K_0$ annotations, obtained via greedy source-distribution refinement, after which full-dataset annotation-balanced exchanges are enabled.
The three image budgets use $K_0=15{,}000$, $30{,}000$, and $60{,}000$, respectively, giving approximately 50.6 annotations per canvas; each count is fixed within its construction run.

\section{Global Objective and Discrete Search}
\label{sec:s_search}

\paragraph{Proof of Proposition 1.}
For a fixed source set and partition structure, all tree-splitting, minimum-size, and scale constraints are affine.
For any $c>0$, $f_c(t)=[\log(c/t)]_+^2$ is convex and nonincreasing on $t>0$: for $0<t<c$, $f_c''(t)=2(1+\log(c/t))/t^2>0$, and both value and first derivative join $0$ continuously at $t=c$.
The term $J_{\mathrm{pad}}$ is a nonnegative sum of positive parts of affine functions.
For $K_0,M_+>0$, all normalizers $K_0$, $M_+$, and $N(W^B+H^B)$ are positive constants; nonnegative loss weights therefore preserve convexity.
With $t_i\in[\epsilon_t,1]$ for fixed $\epsilon_t>0$ and node dimensions bounded by $(W^B,H^B)$, any nonempty feasible set is compact, so continuity guarantees attainment.

\paragraph{Single-source canvas cap.}
We limit the number $n_1$ of canvases containing just one source image so that the layout can combine images across the collection.
Assigning $M\ge N$ images to $N$ nonempty canvases requires at least $\max(0,2N-M)$ such canvases; we permit a fractional slack $\delta$ above this minimum:
\begin{equation}
\label{eq:single}
n_1\le\left\lfloor\max(0,2N-M)+\delta N\right\rfloor.
\end{equation}
We use $\delta=0.05$ and recheck the cap whenever the selected image count changes.

\paragraph{Global scoring.}
For each canvas $j\in\{1,\dots,N\}$, we cache the unnormalized instance resolution sum $A_j$, unrecoverable padding thickness sum $P_j$, and compression sum $C_j$ over source images with valid annotations.
The global continuous objective across the collection is
\begin{equation}
\label{eq:global_score}
J=\frac{\sum_j A_j}{K_0}
+\lambda_{\mathrm{pad}}\frac{\sum_jP_j}{N(W^B+H^B)}
+\lambda_{\mathrm{comp}}\frac{\sum_jC_j}{M_+}.
\end{equation}
We evaluate each solution at $t_i^\star=\min(1,w_i/a_i,h_i/b_i)$ to eliminate scale-variable slack where both resolution penalties vanish.
The caches store unnormalized sums. When an annotation-balanced exchange changes $M_+$ to $M'_+$, only the candidate score uses $M'_+$ to normalize its full compression sum, including cached contributions from unchanged canvases; the current score $J$ is unchanged.
During local search, each proposal re-solves only the one or two affected canvases; unchanged canvas solutions remain feasible because $M_+$ appears only in the objective.
Specifically, let $\mathbf q_j=(A_j,P_j,C_j)$, $\mathbf Q=\sum_j\mathbf q_j$, and let $\mathcal H$ denote the affected canvases.
For a candidate, we update
\[
\mathbf Q'=\mathbf Q+\sum_{j\in\mathcal H}(\mathbf q'_j-\mathbf q_j),
\qquad M'_+=\sum_{i\in\mathcal U'}\ind[K_i>0].
\]
The candidate score is $J'=Q'_A/K_0+\lambda_{\mathrm{pad}}Q'_P/[N(W^B+H^B)]+\lambda_{\mathrm{comp}}Q'_C/M'_+$, whereas $J$ uses the current $\mathbf Q$ and $M_+$. Candidate caches and $M'_+$ replace the current state only upon acceptance.
We accept when $J'<J-10^{-9}$.
After search, we fix the final $M_+$ and re-solve each fixed tree, accepting a feasible refined solution if its contribution is at most the cached contribution plus $10^{-8}$; otherwise, we retain the cached solution.

\paragraph{Proposal distribution.}
Each of the 500 iterations samples a source exchange with probability $0.7$, a cross-canvas swap with probability $0.1$, a cross-canvas relocation with probability $0.1$, or a tree restructuring with probability $0.1$.
These probabilities apply before feasibility checks; an invalid proposal consumes its iteration.
Conditional on a source exchange, the six pairs $(|\mathcal R|,|\mathcal A|)\in\{(1,1),(1,2),(2,1),(2,2),(0,1),(1,0)\}$ each have probability $1/6$.
The last two operations insert or remove one zero-instance image.
For the other operations, removed images are sampled uniformly without replacement; added images come from full-dataset annotation-count buckets whose counts sum to the removed total.
When adding two images, we uniformly sample an available count split $(k,K_{\mathcal R}-k)$, where $K_{\mathcal R}=\sum_{i\in\mathcal R}K_i$.
Each added identity is sampled from its bucket with at most 32 attempts to find an image absent from both the current subset and the other additions; failure rejects the proposal.
We pool the remaining leaves of the affected canvases with the additions and divide this ordered list into groups differing in size by at most one.
A zero-instance insertion chooses one canvas uniformly.

For a swap, we uniformly choose two distinct canvases and one leaf from each, then exchange their source identities.
For a relocation, we uniformly choose two distinct canvases and move a uniformly selected leaf from the first to the second, rejecting moves that empty the first canvas.
For tree restructuring, we uniformly choose one canvas and flip the orientation of a uniformly selected internal node; a single-leaf canvas is skipped.
We then uniformly choose a node with at least one internal child for an associative rotation, $((T_a,T_b),T_c)\leftrightarrow(T_a,(T_b,T_c))$.
If both rotation directions are available, each has probability $1/2$; if only one is available, it is used; if no node is eligible, rotation is skipped.
The orientation flip and rotation belong to the same 10\% branch.

\paragraph{Tree candidates and solver.}
Initial source identities are randomly permuted and split into $N$ groups differing in size by at most one.
For each initial or affected canvas, \textsc{BestTree} tests two fresh balanced trees with vertical and horizontal root splits; each recursively bisects the ordered leaf list, with deeper split directions following the longer side of the provisional region.
For swaps and restructuring, it additionally tests the modified existing tree; restructuring uses a random permutation of its leaves for the two fresh candidates.
New rotation nodes start with split ratio $1/2$, and all candidate dimensions are subsequently optimized.
\textsc{BestTree} returns the feasible solution with the lowest contribution to \cref{eq:global_score}, using the candidate $M'_+$ for every solve.
We use CVXPY with CLARABEL, $t_i\ge10^{-8}$, $d_{\min}=16$, $\tau=16$, $t_{\mathrm{ref}}=0.25$, $\lambda_{\mathrm{pad}}=1$, and $\lambda_{\mathrm{comp}}=0.1$.
Each solution must pass optimal-status, finite-value, primal-residual, and shared-integer-cut minimum-size checks.
All construction hyperparameters are fixed across datasets and budgets.
Algorithm~\ref{alg:joint_search} summarizes the complete joint search and export.

\begin{algorithm*}[t]
\caption{Joint source selection, resolution allocation, and rendering}
\label{alg:joint_search}
\small
\begin{algorithmic}[1]
\Require Source dataset $\mathcal D$; canvas count $N$ and size $(W^B,H^B)$; annotation budget $K_0$; proposal budget $L=500$
\Ensure $N$ static RGB canvases with exactly $K_0$ unique native annotations
\Statex \textbf{Initialize sources and layouts}
\State Compute core rectangles, annotation-count buckets, and source coverage statistics (Appendix~\ref{sec:s_coverage}).
\State Initialize a distinct, coverage-feasible subset $\mathcal U$ with $\sum_{i\in\mathcal U}K_i=K_0$; compute $M_+$.
\State Randomly permute $\mathcal U$ and split it into $N$ nonempty groups of nearly equal size; check the singleton cap.
\State Solve each group with \textsc{BestTree}; cache trees, dimensions, and $\mathbf q_j=(A_j,P_j,C_j)$.
\State $\mathbf Q\gets\sum_j\mathbf q_j$; compute $J$ by \cref{eq:global_score} with $M_+$.
\Statex \textbf{Joint search} (rejected proposals leave the current state unchanged)
\For{$s=1,\ldots,L$}
  \State Sample source exchange / swap / relocation / restructuring with probabilities $(0.7,0.1,0.1,0.1)$.
  \State Apply the proposal rules above to a copy of the current sources and layout; obtain $\mathcal U'$ and affected canvases $\mathcal H$.
  \State If unavailable or violating source uniqueness, exact $K_0$, coverage, nonempty canvases, or the singleton cap: \textbf{continue}.
  \State $M'_+\gets\sum_{i\in\mathcal U'}\ind[K_i>0]$.
  \State Solve only $j\in\mathcal H$ with \textsc{BestTree} using $M'_+$; keep all other canvas solutions cached.
  \State If any solver or integer-partition check fails: \textbf{continue}.
  \State $\mathbf Q'\gets\mathbf Q+\sum_{j\in\mathcal H}(\mathbf q'_j-\mathbf q_j)$.
  \State $J'\gets Q'_A/K_0+\lambda_{\mathrm{pad}}Q'_P/[N(W^B+H^B)]+\lambda_{\mathrm{comp}}Q'_C/M'_+$.
  \If{$J'<J-10^{-9}$}
    \State Accept candidate sources, affected trees and dimensions, caches, $M'_+$, $\mathbf Q'$, and $J'$.
  \EndIf
\EndFor
\Statex \textbf{Refine and export}
\For{each canvas $j=1,\ldots,N$}
  \State Re-solve its fixed tree using the final $M_+$.
  \State Accept a feasible solution if its contribution increases by at most $10^{-8}$; otherwise retain the cached solution.
\EndFor
\State Verify source uniqueness, exact $K_0$, coverage, and singleton cap; freeze shared integer cuts.
\State Recover context, resample source RGB once, and map every native box with realized per-axis scales (Appendix~\ref{sec:s_rendering}).
\State \Return the $N$ canvases and their $K_0$ annotations.
\end{algorithmic}
\end{algorithm*}

\section{Context Geometry and Integer Rendering}
\label{sec:s_rendering}

\paragraph{Placement of the expanded crop.}
For a wider leaf ($\rho_i\ge a_i/b_i$), the continuous expanded dimensions are $W_i^{c\star}=\min(W_i,\rho_i b_i)$ and $H_i^{c\star}=b_i$, positioned horizontally at
\begin{equation}
x_i^{c\star}=\operatorname{clip}\left(
\frac{u_i^-+u_i^+-W_i^{c\star}}{2},0,W_i-W_i^{c\star}\right),
\end{equation}
with vertical span $[v_i^-,v_i^+]$.
Because $W_i^{c\star}\ge a_i=u_i^+-u_i^-$, clamping the centered interval to $[0,W_i-W_i^{c\star}]$ retains $[u_i^-,u_i^+]$ completely; taller leaves ($\rho_i<a_i/b_i$) exchange the two coordinates symmetrically.

\paragraph{Maximal context at fixed continuous scale.}
Any axis-aligned rectangle $C\supseteq U_i$ has width $W^c\ge a_i$ and height $H^c\ge b_i$, so its letterbox scale satisfies $s(C)=\min(w_i/W^c,h_i/H^c)\le s_i^0$.
Requiring $s(C)\ge s_i^0$ enforces $s(C)=s_i^0$ and yields componentwise upper bounds
\begin{equation}
W^c\le\min(W_i,w_i/s_i^0),\quad
H^c\le\min(H_i,h_i/s_i^0).
\end{equation}
Because $C_i^\star$ attains both bounds simultaneously, it maximizes source crop area $W^c H^c$, maximizes rendered RGB area $(s_i^0)^2 W^c H^c$ inside the leaf, and minimizes continuous letterbox padding area while preserving the uncapped scale $s_i^0$.

\paragraph{Strip thickness and padded area.}
Let horizontal and vertical unrecoverable padding thicknesses be
\begin{equation}
p_i^{\mathrm{hor}}=[w_i-\rho_i^{\max}h_i]_+,\qquad
p_i^{\mathrm{ver}}=[h_i-w_i/\rho_i^{\min}]_+.
\end{equation}
Since $\rho_i^{\min}=a_i/H_i\le W_i/b_i=\rho_i^{\max}$, at most one term is positive for any leaf, and $p_i(w_i,h_i)=p_i^{\mathrm{hor}}+p_i^{\mathrm{ver}}$ gives the exact continuous padding-strip thickness remaining after context recovery.
The corresponding continuous padded area is
\begin{equation}
A_i^{\mathrm{pad}}=h_i p_i^{\mathrm{hor}}+w_i p_i^{\mathrm{ver}}.
\end{equation}
Because $A_i^{\mathrm{pad}}$ contains bilinear products of optimization variables, the convex subproblem optimizes linear strip thickness $p_i(w_i,h_i)$, while post-rendering geometry audits measure exact padded pixel fractions on the integer canvases.

\paragraph{Shared integer cuts.}
Given an internal vertical split node of integer width $\bar w_v$, its left child receives width $\operatorname{round}((w_L/w_v)\bar w_v)$ and its right child receives $\bar w_v-\bar w_L$; horizontal splits apply the analogous rule along $\bar h_v$.
Recursive top-down rounding from $(W^B,H^B)$ guarantees a gap-free, nonoverlapping integer tiling, and any proposal producing a leaf dimension below $16$~pixels is rejected.

\paragraph{Integer context expansion and RGB resampling.}
Given realized integer leaf dimensions $(\bar w_i,\bar h_i)$, wider leaves ($\bar w_i/\bar h_i\ge a_i/b_i$) set
\begin{equation}
\bar W_i^c=\min\left(W_i,\max\left(a_i,
\left\lfloor\frac{\bar w_i b_i}{\bar h_i}\right\rfloor\right)\right),
\quad\bar H_i^c=b_i,
\end{equation}
whereas taller leaves ($\bar w_i/\bar h_i<a_i/b_i$) set $\bar W_i^c=a_i$ and
\begin{equation}
\bar H_i^c=\min\left(H_i,\max\left(b_i,
\left\lfloor\frac{\bar h_i a_i}{\bar w_i}\right\rfloor\right)\right),
\end{equation}
using floor rounding so that integer expansion never reduces continuous tight-crop magnification.
After integer centering and clamping to $[0,W_i]\times[0,H_i]$, the rendered RGB dimensions are
\begin{equation}
\begin{split}
\bar s_i&=\min(\bar w_i/\bar W_i^c,\bar h_i/\bar H_i^c),\\
r_i^w&=\max(1,\operatorname{round}(\bar s_i\bar W_i^c)),\\
r_i^h&=\max(1,\operatorname{round}(\bar s_i\bar H_i^c)).
\end{split}
\end{equation}
Each integer crop is resampled once from source RGB via bilinear interpolation to $(r_i^w,r_i^h)$, placed at letterbox offsets $d_i^x=\lfloor(\bar w_i-r_i^w)/2\rfloor$ and $d_i^y=\lfloor(\bar h_i-r_i^h)/2\rfloor$, with remaining border pixels filled with neutral gray ($114$).

\paragraph{Bounding-box mapping.}
Let $(x_i^c,y_i^c)$ be the integer crop origin, $(u_i,v_i)$ the leaf origin on the canvas, and $\alpha_i^x=r_i^w/\bar W_i^c$, $\alpha_i^y=r_i^h/\bar H_i^c$ the realized per-axis resize factors.
Each native box $b=(x,y,\omega,\nu)$ transforms via
\begin{equation}
\label{eq:box}
\widetilde b=\left(\begin{array}{l}
(x-x_i^c)\alpha_i^x+u_i+d_i^x,\\
(y-y_i^c)\alpha_i^y+v_i+d_i^y,\\
\omega\alpha_i^x,\ \nu\alpha_i^y
\end{array}\right).
\end{equation}
This absorbs integer rounding into each axis while preserving annotation IDs, source image IDs, category labels, and all positive floating-point box areas.

Geometry audits record per-axis scales, deviations from integer tight-crop rendering, object-size and annotated-image compression quantiles, and padded pixel fractions.
At every budget, the export satisfies $|\mathcal S|=N$, $\sum_{j=1}^N|\widetilde Y_j|=K_0$, and $B=NW^BH^B$ total pixels: $279{,}752N$ for COCO and $187{,}500N$ for VOC.

%% file: sec/s_sensitivity.tex
\section{Annotation-Budget Sensitivity}
\label{sec:s_k_sweep}

\begin{figure}[t]
\centering
\includegraphics[width=\columnwidth]{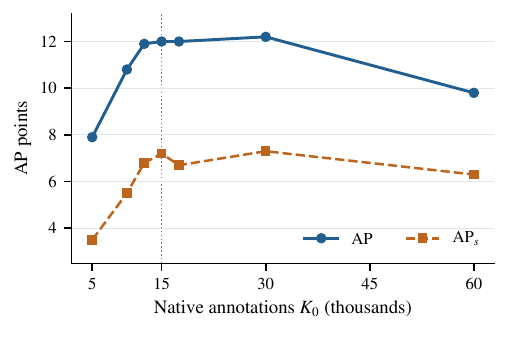}
\caption{\textbf{Annotation-budget sensitivity at fixed resolution.} All runs use 297 canvases of $484\times578$ pixels with Faster R-CNN-R50 (264 epochs). Lines show AP and AP$_s$; the dotted line marks the default $K_0=15$K.}
\label{fig:s_k_sweep}
\end{figure}

\Cref{fig:s_k_sweep} varies the annotation budget $K_0$ from 5K to 60K while keeping the image count ($N=297$) and canvas resolution ($484\times578$) fixed.
Each $K_0$ setting runs coverage initialization and joint source--layout search with the full objective and context recovery, followed by the standard Faster R-CNN-R50 training protocol (batch size 4, two-step gradient accumulation, learning rate 0.02, native augmentations, and hard labels).

AP rises from 7.9 at 5K to 11.9 at 12.5K and remains stable between 12.0 and 12.2 from 15K to 30K, before declining to 9.8 at 60K.
AP$_s$ follows the same trend, increasing from 3.5 at 5K to 7.3 at 30K and dropping to 6.3 at 60K.
These results show that adding supervision within a fixed pixel budget improves detection until excessive compression erodes the visual detail supporting each box, consistent with the canvas--annotation grid in the main paper.

%% file: sec/s_ipd_scaling.tex
\section{Image-Budget Scaling}
\label{sec:s_ipd_scaling}
Table~\ref{tab:ipd_scaling} reports the seven-budget MS COCO sweep with Faster R-CNN-R50 at a fixed canvas resolution of $484\times578$.
Image count $N$ and native annotation budget $K_0$ scale proportionally at approximately 50.6 annotations per canvas.
Overall AP and all three scale-specific metrics improve monotonically with the budget, reaching 34.4 AP and 21.8/37.9/41.5 AP$_s$/AP$_m$/AP$_l$ at 10\% IPD.

\begin{table*}[t]
\centering
\caption{\textbf{MS COCO detection performance across seven image budgets} (\method{}, Faster R-CNN-R50, 264 epochs; AP points, $\uparrow$).}
\label{tab:ipd_scaling}
\small
\setlength{\tabcolsep}{6pt}
\renewcommand{\arraystretch}{1.08}
\begin{tabular*}{\textwidth}{@{\extracolsep{\fill}}rrrrrrrrr@{}}
\toprule
IPD (\%) & Images ($N$) & Annotations ($K_0$) & AP & AP$_{50}$ & AP$_{75}$ & AP$_s$ & AP$_m$ & AP$_l$ \\
\midrule
0.125 & 149 & 7,500 & 8.6 & 18.5 & 7.0 & 4.7 & 10.4 & 9.7 \\
0.25 & 297 & 15,000 & 12.0 & 24.1 & 11.0 & 7.2 & 14.3 & 13.7 \\
0.50 & 593 & 30,000 & 15.2 & 29.1 & 14.3 & 9.3 & 17.6 & 17.2 \\
1.00 & 1,185 & 60,000 & 18.5 & 33.4 & 18.5 & 10.8 & 20.8 & 21.3 \\
2.00 & 2,370 & 120,000 & 22.5 & 38.7 & 23.2 & 13.6 & 25.4 & 25.3 \\
5.00 & 5,925 & 300,000 & 29.0 & 47.3 & 30.3 & 18.2 & 32.2 & 33.5 \\
10.00 & 11,850 & 600,000 & 34.4 & 54.2 & 36.8 & 21.8 & 37.9 & 41.5 \\
\bottomrule
\end{tabular*}
\end{table*}

%% file: sec/s_qualitative.tex
\section{Visualizing the Distilled Images}
\label{sec:s_qualitative}
\begin{figure}[t]
\centering
\includegraphics[width=\columnwidth]{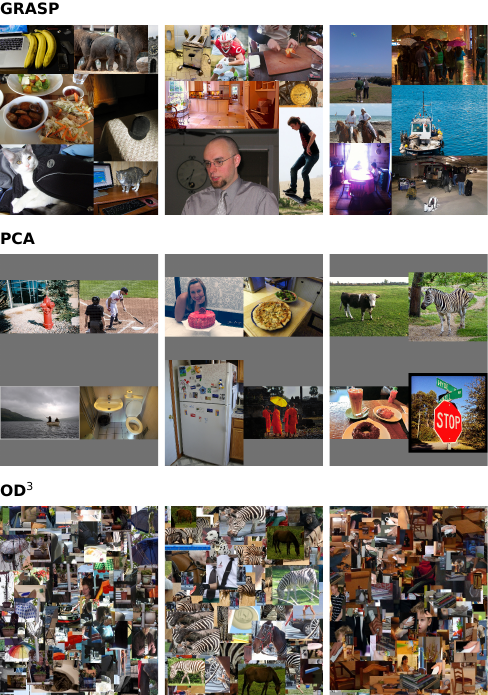}
\caption{\textbf{Distilled RGB images from three construction methods.} Each row shows three canvases at the 0.25\% MS COCO budget: \method{}, PCA, and \odthree{}, from top to bottom. Images retain their native aspect ratios and are shown without box overlays.}
\label{fig:s_qualitative}
\end{figure}

\Cref{fig:s_qualitative} compares distilled RGB canvases from \method{}, PCA, and \odthree{} at the 0.25\% MS COCO budget.
\method{} packs multi-object scene crops of adaptive sizes while preserving the spatial layout within each crop.
In contrast, PCA arranges four crops on a uniform grid, and \odthree{} densely tiles isolated instance crops.
The PCA canvases are produced by our detection adaptation, and the \odthree{} canvases are taken from the official release.\footnote{\url{https://huggingface.co/datasets/VILA-Lab/OD3}}

